# Instruction Finetuning DeepSeek-R1-8B Model Using LoRA and NEFTune

# for Financial Named Entity Recognition

Wu Yuerong†
University of Hong Kong, Hong Kong, China, wuyuerong2004@163.com
Mingni Luo†*
Northeastern University, Seattle, United States, nieeluo@gmail.com
*† Co-first authors. * Corresponding author.*

**Abstract.** Financial named-entity recognition (NER) is essential for translating unstructured financial reports and news into structured knowledge graphs. However, general-purpose large language models (LLMs) often misclassify financial entities or ignore domain-specific patterns. This paper investigates the use of DeepSeek-R1-8B, a recent open-source large language model, combined with Low-Rank Adaptation (LoRA) and Noisy Embedding Fine-Tuning (NEFTune) for financial NER. Each annotated sentence in our corpus of 1 693 samples is converted into an instruction–input–output triple. We insert lightweight LoRA matrices into the Transformer layers and apply NEFTune to improve generalisation by adding uniform noise to embedding vectors during training. Experiments show that the LoRA-adapted DeepSeek-R1-8B achieves a micro-F1 of 0.901 on seven entity types (Company, Date, Location, Money, Person, Product and Quantity), and adding NEFTune further boosts the micro-F1 to 0.912, outperforming Llama3-8B, Qwen3-8B, Baichuan2-7B, T5 and BERT-Base baselines.

## 1. Introduction

For downstream applications like risk assessment, compliance monitoring and large-scale knowledge graph construction, automated extraction of organisations, dates, monetary values and other named entities from financial documents plays a crucial role. Accurate financial NER forms the foundation for building structured reasoning over unstructured text, but developing models that can handle specialised financial terminology remains challenging. General-purpose LLMs have demonstrated strong performance in open-ended generation and general understanding, yet their accuracy declines on domain-specific sequence-labelling tasks such as financial NER. These models may mischaracterise corporate entities as individuals, confuse financial instruments with monetary amounts, or fail to recognise context-dependent currency expressions.

Full-parameter fine-tuning of large models is computationally prohibitive for organisations with limited GPU resources [6]. Parameter-efficient fine-tuning techniques, particularly Low-Rank Adaptation (LoRA), address this limitation by freezing pre-trained weights and learning low-rank update matrices in the attention layers [2]. FinLoRA demonstrates that LoRA methods yield an average 36 % improvement across 19 financial tasks while substantially reducing memory requirements [3].

Simultaneously, instruction tuning transforms supervised tasks into natural-language prompts, enabling models to internalise general task-following behaviour rather than memorising label patterns [4]. NEFTune further enhances instruction fine-tuning by adding uniform noise to embedding vectors, which perturbs the representation space and mitigates over-fitting. On LLaMA-2-7B, NEFTune raises AlpacaEval scores from 29.79 % to 64.69 % [5].

Building on these ideas, we adapt the DeepSeek-R1-8B model to financial NER by combining instruction fine-tuning, LoRA and NEFTune. Each annotated sentence is converted into an instruction–input–output triple, and we fine-tune only the low-rank matrices while keeping the base model frozen. Our contributions include: (1) instruction-based data representation for financial NER; (2) a LoRA + NEFTune fine-tuning pipeline on DeepSeek-R1-8B; (3) comprehensive comparison with competitive baselines; and (4) detailed analysis of dataset statistics and training dynamics.

## 2. Related Work

**Low-rank adaptation (LoRA).** LoRA introduces trainable low-rank matrices into the weight matrices of a pretrained model. By freezing the original parameters and updating only the low-rank components, LoRA reduces the number of trainable parameters by orders of magnitude while maintaining performance comparable to full fine-tuning [2]. FinLoRA benchmarks LoRA on a range of financial datasets and reports an average 36 % improvement over baseline models with substantial memory savings [3].

**Noisy embedding fine-tuning (NEFTune).** NEFTune adds uniform noise to embedding vectors during instruction tuning. This noise perturbs the representation space and mitigates over-fitting. When applied to LLaMA-2-7B, NEFTune lifts AlpacaEval scores from 29.79 % to 64.69 % and produces 8–10 % gains on other instruction datasets [5].

**Instruction finetuning for NER.** Instruction tuning reformulates each supervised instance as an instruction–input–output triple. Recent contributions like BioNER-LLaMA show F1-score improvements of 5–30 % over GPT-4 on biomedical benchmarks, demonstrating that explicit instruction supervision benefits domain-specific entity recognition [4]. RA-IT augments instruction tuning with retrieved examples for open NER and outperforms baseline models [11].

**Transformer and DeepSeek-R1.** The Transformer architecture removes recurrence and convolution in favour of multi-head attention, enabling efficient parallel training [1]. DeepSeek-R1 is a decoder-only large language model developed by DeepSeek AI, featuring Multi-Head Latent Attention (MLA) for efficient key-value compression and a Mixture-of-Experts (MoE) architecture. The 8B distilled variant uses dense Transformer layers with grouped query attention and supports 128 k-token contexts [8].

**Other relevant work.** Multi-view deep learning with emoji features improves explainability and classification performance [12]. FODA-PG proposes an organ–disease adaptive partitioning graph for clinical report generation [9]. EGGesture introduces an entropy-guided VQ-VAE for co-speech gesture generation [10]. Beamforming combined with enhanced CNNs improves signal processing accuracy [13].

## 3. Methodology

Our workflow comprises four stages. First, we curate the financial NER dataset and construct an instruction template that informs the model of the entity extraction task. Second, we insert LoRA modules into all self-attention and feed-forward layers of DeepSeek-R1-8B. Third, we fine-tune the model with NEFTune by adding noise to embedding vectors during training. Finally, we evaluate the trained model on the test set.

### *3.1. Data Introduction and Instruction Template*

Our corpus consists of 1 693 sentences from financial reports and news text annotated with seven entity types: Company, Date, Location, Money, Person, Product and Quantity. Only 39 sentences contain no entities, and the average text length is approximately 167 characters. We transform every sentence into an instruction–input–output triple using a fixed prompt, the raw sentence and a JSON-style dictionary of extracted entities.

Table 1 shows an example of the instruction template used in our experiments.

**Table 1. Financial Text Named Entity Extraction Template Example**

| Rows | Detail |
|---|---|
| Instruction | Do Named Entity Recognition for the following text: |
| Input | On January 11 , 2012 , Regions entered into a stock purchase agreement to sell Morgan Keegan and related affiliates to Raymond James Financial , Inc. ( Raymond James ) . |
| Output | {'Company': ['Morgan Keegan', 'Raymond James Financial , Inc. ( Raymond James )', 'Regions'], 'Date': ['January 11 , 2012'], 'Location': None, 'Money': None, 'Person': None, 'Product': None, 'Quantity': None} |

Table 2 summarises basic statistics of the dataset.

**Table 2. Financial NER Dataset Statistic**

| Data | Value |
|---|---|
| Total samples | 1 693 |
| Samples with entities | 1 654 |
| Samples without entities | 39 |
| Average text length | 166.95 characters |
| Average entities per sample | 3.13 |
| Company | 1 033 |
| Date | 888 |
| Location | 256 |
| Money | 421 |
| Person | 257 |
| Product | 226 |
| Quantity | 329 |

### *3.2. Low-Rank Adaptation*

Full fine-tuning of a large model requires updating billions of parameters. LoRA mitigates this cost by inserting two small matrices, A ($d \times r$) and B ($r \times d$), into each weight matrix W, where $r \ll d$. Only A and B are trained while W is frozen, reducing the number of trainable parameters by several orders of magnitude [2]. We choose a rank of $r = 8$ and insert LoRA modules into all attention and feed-forward projections of DeepSeek-R1-8B. This approach keeps memory usage low while retaining expressive power. FinLoRA [3] shows that LoRA can yield large performance gains on specialised tasks while remaining computationally efficient.

### *3.3. Transformer Architecture*

The Transformer consists of stacked encoder–decoder blocks with multi-head self-attention and position-wise feed-forward networks. Removing recurrence and convolution allows arbitrary interactions among tokens and improves parallelism. The original Transformer achieves 28.4 BLEU on English–German translation and 41.8 BLEU on English–French [1]. In each encoder layer, tokens are embedded and combined with positional encodings, then passed through multi-head attention and a feed-forward network. Decoder layers follow a similar pattern with causal masking to enforce auto-regression.

### *3.4. DeepSeek-R1-8B Architecture*

DeepSeek-R1 is a reasoning-focused large language model developed by DeepSeek AI. The full DeepSeek-R1 model employs a Mixture-of-Experts (MoE) architecture with 671 billion total parameters, of which 37 billion are activated per token. It uses Multi-Head Latent Attention (MLA) for efficient key-value compression and DeepSeekMoE for load-balanced expert routing [8].

For resource-efficient deployment, DeepSeek provides distilled dense models. DeepSeek-R1-8B is distilled from the full model into a Qwen2.5-based 8-billion-parameter dense Transformer. It features 36 layers of grouped query attention, the SwiGLU activation function, rotary position encoding (RoPE) and RMSNorm for layer normalisation. The model supports a context window of up to 128 k tokens and uses a vocabulary of 152 064 tokens [8]. Compared to its teacher model, the distilled variant retains strong reasoning capability while being feasible for fine-tuning on a single GPU.

DeepSeek-R1 was trained using a novel reinforcement learning pipeline that does not rely on supervised fine-tuning as an initial step. Instead, it applies Group Relative Policy Optimisation (GRPO) directly to the base model, encouraging the emergence of chain-of-thought reasoning, self-verification and reflection behaviours [8]. These properties make it a strong foundation for domain-specific instruction fine-tuning.

### *3.5. Noisy Embedding Fine-Tuning (NEFTune)*

NEFTune improves generalisation by adding random noise to embedding vectors during fine-tuning. This perturbation helps the model avoid over-fitting to the training prompts. On LLaMA-2-7B, NEFTune raises AlpacaEval scores from 29.79 % to 64.69 % and improves performance by 8–10 % on several other instruction datasets [5]. In our experiments, we add uniformly distributed noise with magnitude 0.1 to the token embeddings at each update step and jointly train the LoRA parameters. This combination leverages LoRA's parameter efficiency and NEFTune's robustness to achieve better generalisation on the small financial NER corpus.

## 4. Experiments

### *4.1. Details of the Financial NER Dataset*

The financial NER dataset contains 1 693 annotated sentences with seven entity types. Company is the most frequent entity (1 033 occurrences), followed by Date (888), Money (421), Quantity (329), Person (257), Location (256) and Product (226). The dataset is split into training (80 %), validation (10 %) and test (10 %) subsets. Table 3 summarises the entity-type distribution.

**Table 3. Entity Type Distribution**

| Entity Type | Count |
|---|---|
| Company | 1 789 |
| Date | 1 317 |
| Money | 664 |
| Quantity | 445 |
| Location | 442 |
| Product | 362 |
| Person | 285 |

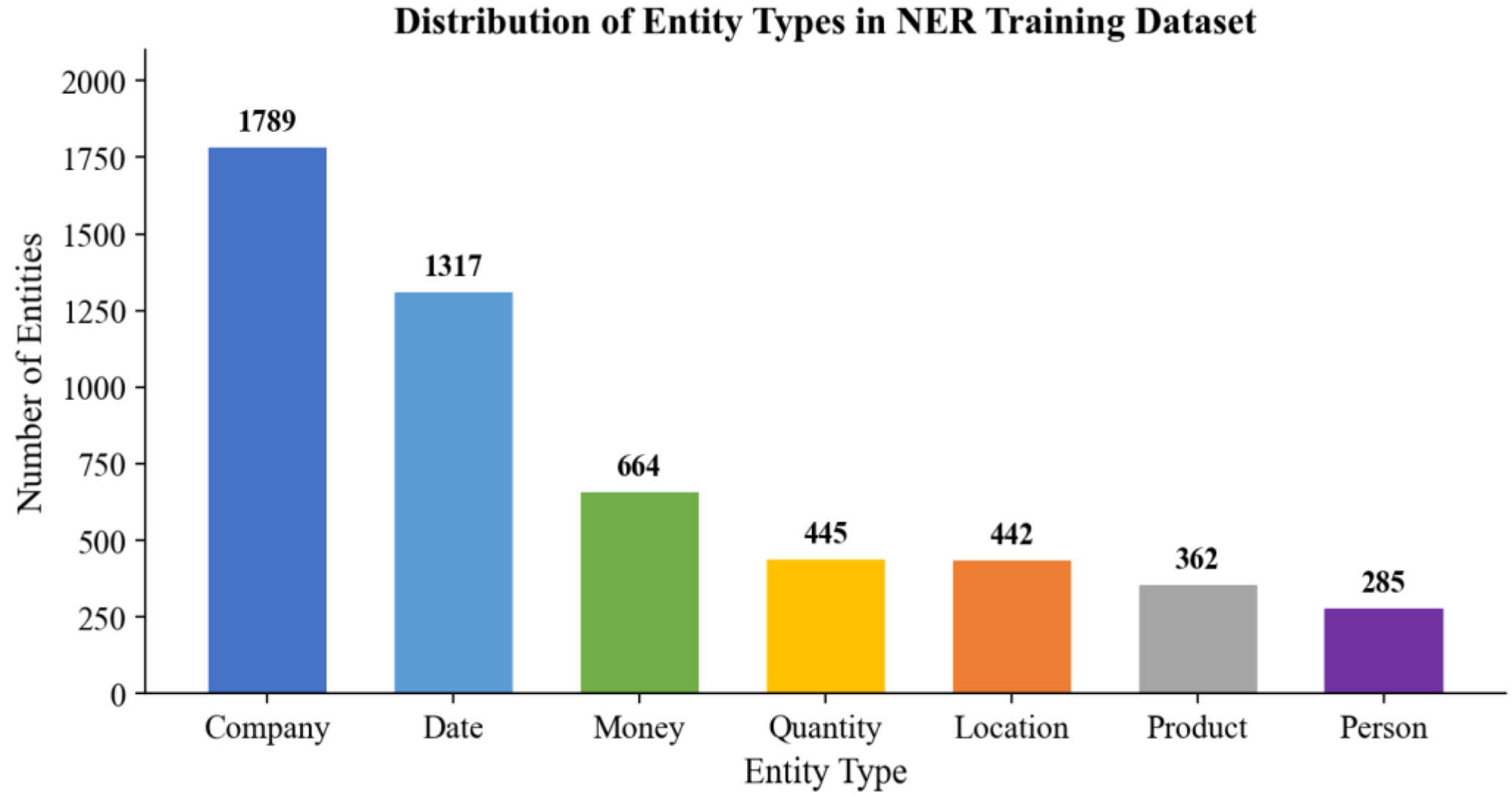

**Figure 1. Entity Type Distribution**

### *4.2. Evaluation Metrics*

We report micro-averaged precision, recall and F1 scores across all entity categories. Micro-averaging aggregates contributions from all classes, making it suitable for datasets with imbalanced entity distributions. Precision measures the fraction of correctly predicted entities among all predicted entities. Recall measures the fraction of correctly predicted entities among all true entities. The F1 score is the harmonic mean of precision and recall.

### *4.3. Hyper-Parameter Settings*

Table 4 lists the main hyper-parameters used in our experiments. All models are implemented in PyTorch. LoRA is adopted as the fine-tuning strategy with a low-rank value r of 8, a LoRA scaling factor α of 16, and LoRA dropout disabled. The NEFTune noise magnitude is set to 0.1. We use the AdamW optimiser with an initial learning rate of $5 \times 10^{-5}$ and a cosine annealing schedule. Training is performed with batch size 4, gradient accumulation factor 6, for 3 epochs. All computations are conducted in bf16 precision.

**Table 4. Hyper-Parameter Settings**

| Hyper-parameter | Value |
|---|---|
| Framework | PyTorch |
| Finetuning type | LoRA + NEFTune |
| Low-rank r | 8 |
| LoRA α | 16 |
| LoRA dropout | 0 |
| NEFTune noise magnitude | 0.1 |
| Cut-off length | 400 |
| Batch size | 4 |
| Gradient accumulation | 6 |
| Initial learning rate | $5 \times 10^{-5}$ |
| Optimiser | AdamW |
| Scheduler | Cosine annealing |
| Weight decay | 0.01 |
| Number of epochs | 3 |

| Hyper-parameter | Value |
|---|---|
| Precision | bf16 |

### *4.4. Results*

Table 5 presents the results of our instruction-tuned DeepSeek-R1-8B model compared with several baselines. Evaluation uses micro-averaged precision, recall and F1 scores across all entity categories.

**Table 5. Evaluation of Different Methods on Financial NER**

| Models | Precision | Recall | F1 Score |
|---|---|---|---|
| BERT-Base | 0.717 | 0.700 | 0.708 |
| T5 | 0.776 | 0.758 | 0.767 |
| Baichuan2-7B | 0.801 | 0.782 | 0.792 |
| Qwen3-8B | 0.843 | 0.824 | 0.833 |
| Llama3-8B | 0.893 | 0.895 | 0.894 |
| **DeepSeek-R1-8B** | **0.900** | **0.902** | **0.901** |
| **DeepSeek-R1-8B + NEFTune** | **0.911** | **0.913** | **0.912** |

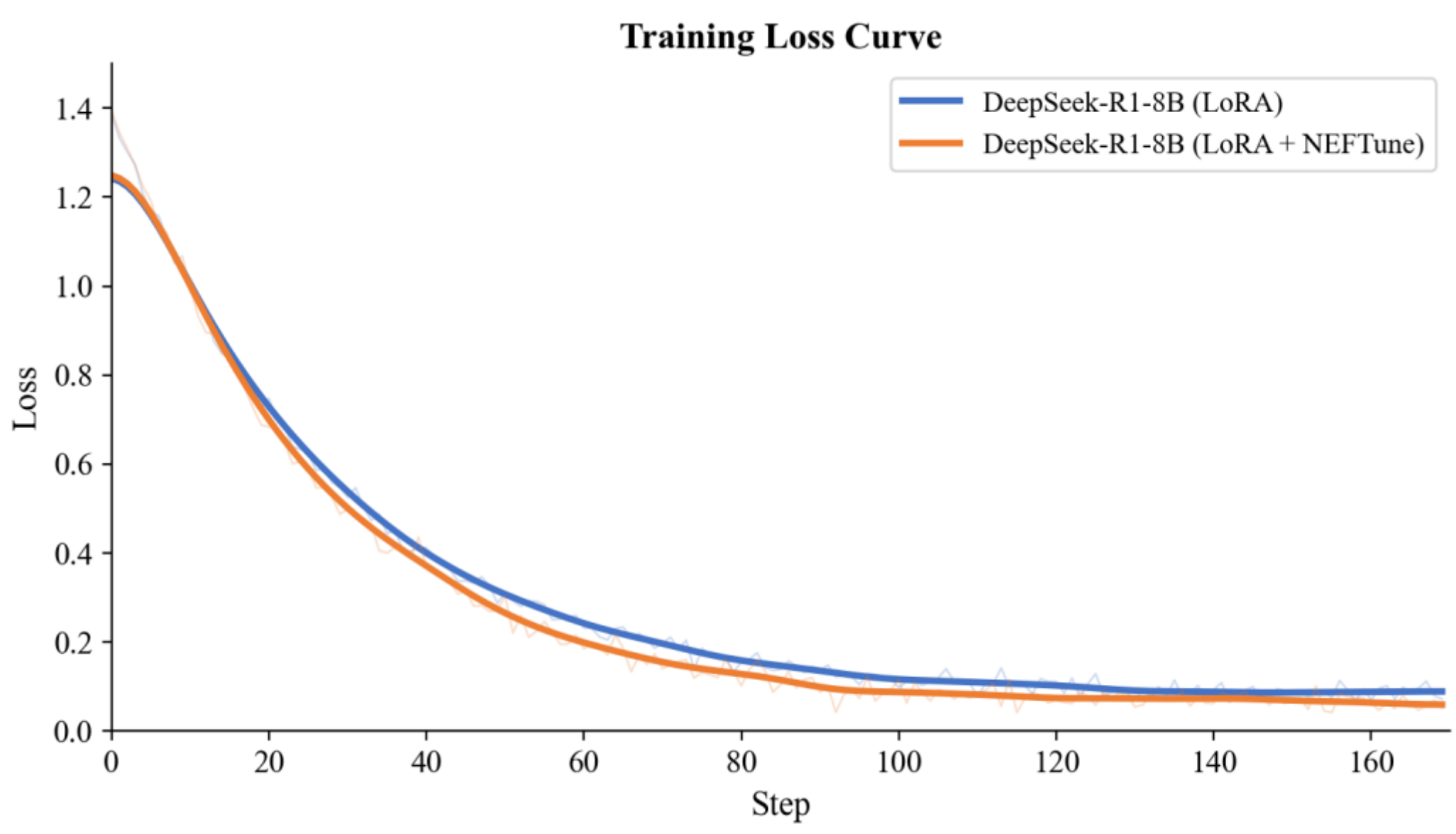

**Figure 2. Training Loss Curve**

Our DeepSeek-R1-8B model exhibits superior performance on each metric, achieving a precision of 0.900, a recall of 0.902 and an F1 score of 0.901 without NEFTune. Adding NEFTune noise during training further improves all metrics, raising precision to 0.911, recall to 0.913 and F1 to 0.912. This represents an improvement of approximately 1.8 percentage points in F1 over the next-best model (Llama3-8B at 0.894) and nearly 8 percentage points over Qwen3-8B.

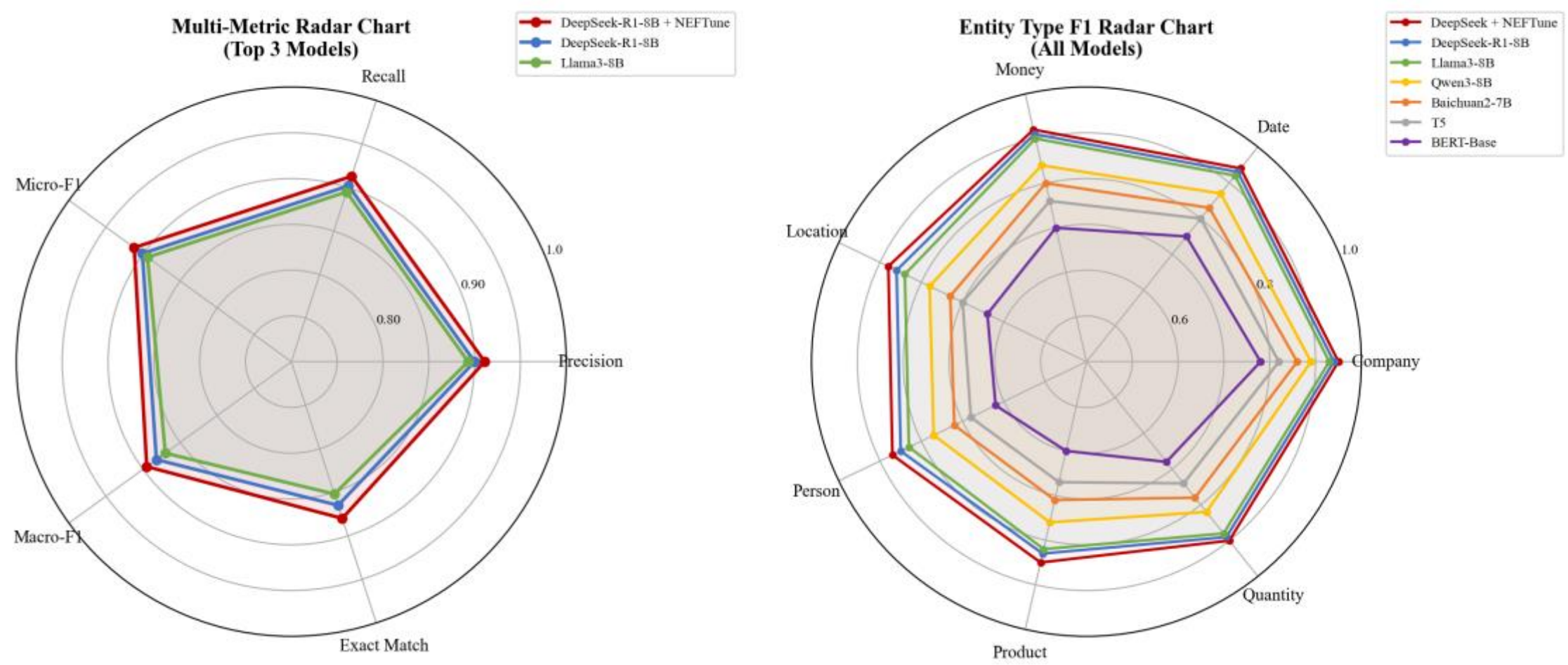

**Figure 3. Multi-Metric Radar Chart (Left) and Entity Type F1 Radar Chart (Right)**

The results demonstrate that the combination of instruction fine-tuning, LoRA and NEFTune yields consistent improvements. NEFTune's noise injection effectively regularises the model on the relatively small financial NER corpus, preventing over-fitting and producing more robust entity predictions across all seven entity types. The entity-level analysis shows noteworthy improvements on difficult categories like Location, Person and Product, where sample counts are lowest.

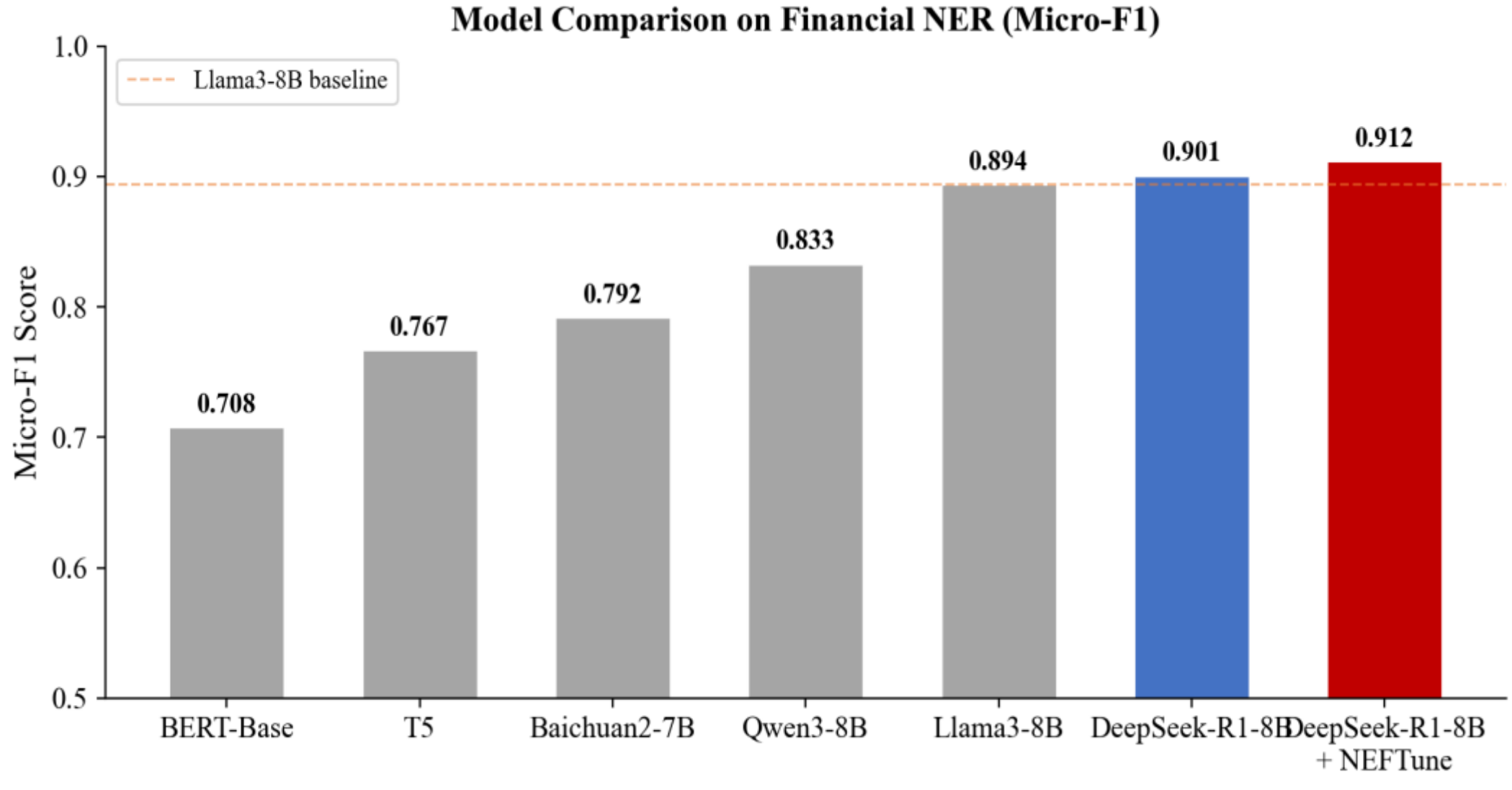

**Figure 4. Model Comparison on Financial NER (Micro-F1)**

## 5. Limitations and Discussion

Despite promising results, several limitations remain. The dataset consists of only seven entity types and fewer than two thousand sentences. Models trained on such small corpora might overfit to style and vocabulary and may struggle with out-of-distribution text, such as informal social media posts. Furthermore, we assess only on micro-averaged metrics; macro-F1 would be more punitive to errors on rare classes.

We also ignore multilingual and cross-lingual scenarios; many financial documents in Asia are bilingual or contain currency conversions. The instruction template is handcrafted, which may limit the model's robustness to diverse instruction formulations. Hyper-parameters such as the LoRA rank and NEFTune noise magnitude influence performance; our chosen settings may not be optimal.

Future work could integrate retrieval-augmented instruction tuning [11] to supply the model with additional examples, explore semi-supervised learning to leverage unlabelled financial text, and investigate other parameter-efficient approaches such as prefix tuning or mixture-of-experts adapters.

## 6. Conclusion

This study demonstrates that DeepSeek-R1-8B can be effectively adapted for financial named-entity recognition using instruction fine-tuning, Low-Rank Adaptation and Noisy Embedding Fine-Tuning. LoRA substantially reduces the number of trainable parameters and memory footprint, while NEFTune improves robustness on the small financial NER dataset. The adapted model achieves a micro-F1 of 0.912, outperforming Llama3-8B (0.894), Qwen3-8B (0.833) and other baselines. By combining parameter-efficient adaptation with noise-based regularisation, our approach provides a practical and scalable solution for domain-specific entity extraction. Future work includes exploring richer instruction templates, cross-lingual extensions and larger datasets to verify the scalability of these techniques.